\documentclass[letterpaper, 10 pt, conference]{ieeeconf}
\IEEEoverridecommandlockouts
\usepackage{url}

\usepackage{tabu}
\usepackage{booktabs} 
\usepackage{multirow}
\usepackage{multicol}
\usepackage{adjustbox}

\usepackage{xcolor}
\usepackage{xspace}
\usepackage{ifthen}

\usepackage{graphicx}
\usepackage[font=small]{caption}
\usepackage{tikz}
\usetikzlibrary{positioning}

\let\labelindent\relax 

\usepackage{autolabtools}
\usepackage{amsmath,amssymb,amsfonts}
\usepackage{dsfont}
\usepackage{color}
\usepackage{verbatim}
\usepackage{subcaption}
\usepackage{balance}
\usepackage{gensymb}
\usepackage{siunitx} 
\usepackage{algorithm}
\usepackage{algorithmicx}
\usepackage{algpseudocode}
\usepackage{alphalph}
\usepackage{comment}
\usepackage{soul}
\usepackage[shortlabels]{enumitem}

\graphicspath{{figures/}}
\newcommand{\ba}{\mathbf{a}}

\newcommand{\bs}{\mathbf{s}}

\newcommand{\eg}{e.g., }

\newcommand{\R}{\mathbb{R}}

\newcommand{\mO}{\mathcal{O}}

\allowdisplaybreaks[4]

\newcommand{\remark}[3]{\hidable{{\color{#2}[#1: #3]}}}

\definecolor{britishracinggreen}{rgb}{0.23, 0.53, 0.19}

\newcommand{\daniel}[1]{\remark{Daniel}{blue}{#1}}

\definecolor{navy}{rgb}{0,0,0.5}

\newcommand{\bagging}{AutoBag\xspace}

\usepackage[font={small}]{caption}
\def\tablename{Table}

\usepackage[backend=biber,
            hyperref=true,
            url=false,
            isbn=false,
            doi=false,
            backref=false,
            style=ieee,
            natbib=true,
            mincitenames=1,
            maxcitenames=1,
            citestyle=numeric-comp,
            sorting=nyt,
            block=none,
            maxbibnames=99]{biblatex}
\usepackage{hyperref}

\title{\LARGE \bf
AutoBag: Learning to Open Plastic Bags and Insert Objects
}

\author{Lawrence Yunliang Chen$^1$, Baiyu Shi$^1$, Daniel Seita$^2$, \\ Richard Cheng$^3$, Thomas Kollar$^3$, David Held$^2$, Ken Goldberg$^1$
\thanks{$^{1}$The AUTOLab at UC Berkeley (automation.berkeley.edu).}%
\thanks{$^{2}$The Robotics Institute at Carnegie Mellon University.}
\thanks{$^{3}$Toyota Research Institute, Los Altos, USA.}
\thanks{Correspondence to: {\tt\scriptsize yunliang.chen@berkeley.edu}}
}

\begin{document}

\maketitle

\thispagestyle{empty}
\pagestyle{empty}

\begin{abstract}
Thin plastic bags are ubiquitous in retail stores, healthcare, food handling, recycling, homes, and school lunchrooms. 
They are challenging both for perception (due to specularities and occlusions) and for manipulation (due to the dynamics of their 3D deformable structure). 
We formulate the task of ``bagging:'' manipulating common plastic shopping bags with two handles from an unstructured initial state to an open state where at least one solid object can be inserted into the bag and lifted for transport. 
We propose a self-supervised learning framework where a dual-arm robot learns to recognize the handles and rim of plastic bags using UV-fluorescent markings; at execution time, the robot does not use UV markings or UV light. We propose the \bagging algorithm, where the robot uses the learned perception model to open a plastic bag through iterative manipulation. 
We present novel metrics to evaluate the quality of a bag state and new motion primitives for reorienting and opening bags based on visual observations. 
In physical experiments, a YuMi robot using \bagging is able to open bags and achieve a success rate of 16/30 for inserting at least one item across a variety of initial bag configurations. Supplementary material is available at \url{https://sites.google.com/view/autobag}.
\end{abstract}

\section{Introduction}
Opening thin plastic bags and then inserting objects for efficient transport is a useful skill for tasks such as grocery shopping, cleaning, recycling, and packing. However, this is very difficult for robots. Deformable thin objects are challenging to manipulate due to their infinite-dimensional state spaces and nonlinear dynamics, and while there has been much prior work on deformable object manipulation, most focus on 1D linear objects such as ropes~\cite{state_estimation_LDO,priya_2020,rope_untangling_2013}, cables~\cite{tactile_cable_2020}, and elastic beams~\cite{duenser2018interactive,zimmermanndynamic}, or 2D objects such as fabrics~\cite{maitin2010cloth,fabricflownet,yan_fabrics_latent_2020,seita_fabrics_2020}, gauze~\cite{thananjeyan2017multilateral}, and paper~\cite{origami_2015,deformation_page_turning_2021}. The 3D structure and elastoplastic material of plastic bags present numerous challenges. 
Such bags are extremely lightweight, and moving parts of a bag with one gripper often results in the entire bag moving without much change in the opening. Thus, opening a thin plastic bag requires coordination among multiple contacts. 
Furthermore, many plastic bags are reflective, translucent, or transparent, making perception challenging. 

\begin{figure}[t]
\center
\includegraphics[width=0.49\textwidth]{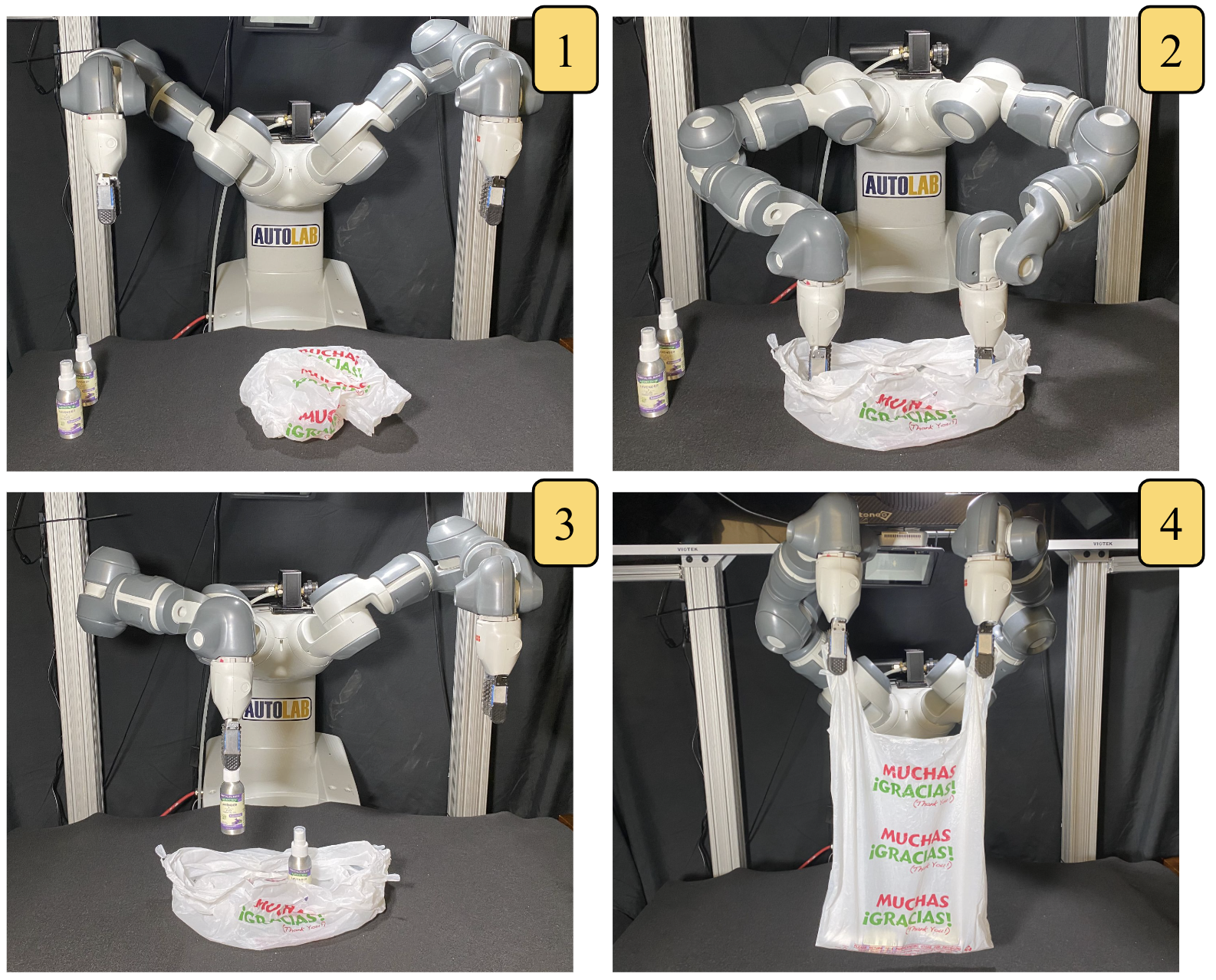}
\caption{
\bagging. (1) Initial highly unstructured and deformed bag. (2) After a sequence of manipulation steps, the robot orients the bag upward and opens the bag. (3) The robot inserts 2 items into the bag. (4) The robot lifts the bag filled with the inserted items, so it is ready for transport.
}
\vspace{-15pt}
\label{fig:teaser}
\end{figure}

In this work, we formulate ``\emph{bagging}''---manipulating a plastic bag from an unstructured initial state so that a robot can open it, insert solid objects into it, and then lift it for transport.
We use overhead RGB images for perception and a bimanual robot. 
We propose a novel pipeline for bagging that trains a perception model to segment the bag rim and handles through self-supervised data collection. This involves the robot systematically exploring the bag state space by manipulating a bag annotated with ultraviolet (UV) labels~\cite{LUV_2022}. At test time, we deploy the learned segmentation model on bags without UV labels and evaluate the bag opening using novel metrics. We present a novel algorithm, \bagging, for opening and inserting items into bags. See Figure~\ref{fig:teaser}.

This paper makes 6 contributions:
\begin{enumerate}[leftmargin=*]
    \item A novel problem formulation for ``bagging;'' 
    \item A novel set of primitive actions for manipulating bags including shaking, compressing, flipping, and dilation;
    \item A self-supervised data collection process where a robot efficiently explores its state space to manipulate bags into diverse configurations to enable recognizing the handles and rim of bags from UV-fluorescent markings;
    \item Two metrics that quantify the opening of the bag based on the convex hull area and elongation; 
    \item The \bagging algorithm for bagging; 
    \item An implemented system with experimental results achieving a success rate of 53.3\% for inserting at least 1 object over 30 physical trials. 
\end{enumerate}

\section{Related Work}

\subsection{Deformable Object Manipulation}

Deformable object manipulation remains challenging for robots~\cite{manip_deformable_survey_2018,grasp_centered_survey_2019,2021_survey_defs}. Typical reasons include the complex dynamics and the infinite set of possible configurations. As a rough categorization, deformable object manipulation can be divided into tasks that involve 1D, 2D, or 3D objects.

Manipulation of 1D deformable objects refers to manipulation of items such as cables~\cite{wire_insertion_1996,wire_insertion_1997,tactile_cable_2020,zhu_sliding_cables_2019,PRC_2022}, ropes~\cite{nair_rope_2017,harry_rope_2021,wang_visual_planning_2019}, and other items which can largely be defined by a single linear component. These are used in tasks such as knotting~\cite{knot_planning_2003,case_study_knots_1991} or untangling~\cite{untanglingLongCables2022,grannen2020untangling}.
Manipulation of 2D objects refers to items such as clothing and fabrics, as studied in recent work on fabric smoothing~\cite{flinging_2022,bodies_uncovered_2022,fabric_vsf_2020,seita-bedmaking,ha2021flingbot,fabricflownet,VCD_cloth,lerrel_2020,gdoom2021}, which often measuring quality using coverage. A smooth fabric with high coverage may make it easier to later do folding, another canonical task explored in prior work~\cite{sim2real_deform_2018,descriptors_fabrics_2021,folding_fabric_fcn_2020,speedfolding_2022}. Assistive dressing~\cite{ra-l_dressing_2018,assistive_gym_2020} encompasses a third set of tasks utilizing 2D deformables.  

Manipulation of 3D deformable objects adds another dimension. One type of 3D deformable manipulation involves volumetric 3D objects, including plush toys, sponges, and dough. Prior work has studied manipulation of these items to target configurations~\cite{ACID2022,Qi_dough_2022,matl2021Deformable,PASTA_2022}. A second type of 3D deformable manipulation refers to objects typically held in containers, as in manipulation of liquids~\cite{visual_closed_loop_liquids_2017} and granular media~\cite{schenck_2017,samuel_clarke_2018,matl2020inferring}, which may require scooping policies~\cite{grannen_scooping}. 
Other references to 3D deformable manipulation refer to thin surfaces arranged in complex 3D patterns, such as plastic bags, which is the main focus of this work.

\subsection{Manipulating Deformable Bags}

One prior direction in robot manipulation of bags is on the mechanical design of robots suitable for grasping~\cite{grasping_sacks_2005} or unloading~\cite{unloading_sacks_2008} large sacks. Another direction has provided insights on manipulating knotted bags~\cite{ayanna_1996,ayanna_2000} or bags in highly constrained setups, such as with work on closing ziplock bags~\cite{contour_ziplock_2018} or using fully opened, stable paper grocery bags~\cite{checkout_robot_2011}.
Recently, DextAIRity~\cite{dextairity2022} used air to efficiently expand bags. They used a setup with three UR5 robots, where two grip the bag and the third manipulated a leaf blower in free space. 
In this work, we consider a bimanual robot with standard parallel-jaw end-effectors for a highly deformable plastic bag with handles, which the robot has to grasp, open, and then insert items inside for transport.

Seita~et~al.~\cite{seita_bags_2021} proposed several deformable object manipulation benchmark tasks in simulation that include a similar problem setup of opening and inserting items into a bag, and then lifting and moving the bag. 
Transporter Networks~\cite{zeng_transporters_2020} are proposed but only evaluated in simulation~\cite{coumans2019}. Similarly, Weng~et~al.~\cite{graph_based_interaction_2021} study modeling and interacting with bags purely in simulation. A large sim-to-real gap in both visual and dynamic properties still needs to be overcome for a policy trained in simulation to transfer to real. 
Some research with deformable bags~\cite{seita_bags_iros_2021,bahety2022bag} assumes that the bag starts with its rim facing upwards and the bag wide open to simplify item insertion and instead focus on the object rearrangement and the bag lifting steps. In this work, we allow bags to start from unstructured configurations and tackle the challenge of orienting and opening the bag.

In recent work, Gao~et~al.~\cite{bagKnotting2022} proposed an algorithm for tying the handles of deformable plastic bags. 
To facilitate tying, they fill in bags beforehand with items which expand the bags and make their handles more likely to be upright and exposed. 
Instead of tying filled bags, we study a different problem setting of opening and inserting items into bags, where the bags begin empty and in unstructured states.

\begin{figure}[t]
\center
\includegraphics[width=0.49\textwidth]{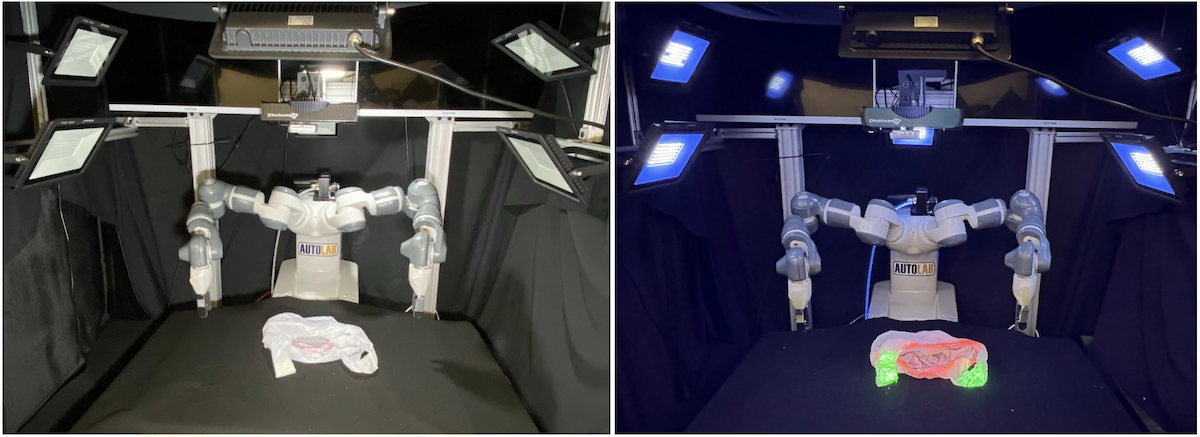}
\caption{
The physical setup with the ABB YuMi. During training, we use 6 UV LED lights and 1 regular LED light. We use a Realsense RGBD camera placed overhead the robot. \textbf{Left}: The workspace under regular lighting. \textbf{Right}: The workspace and a painted bag under UV lighting. The painted rim and handle regions of the bag look normal under regular lighting but glow under UV lights. The UV lights are only used during training, not during execution time.
}
\vspace{-5pt}
\label{fig:camera_setup}
\end{figure}

\begin{figure}[t]
\center
\includegraphics[width=0.49\textwidth]{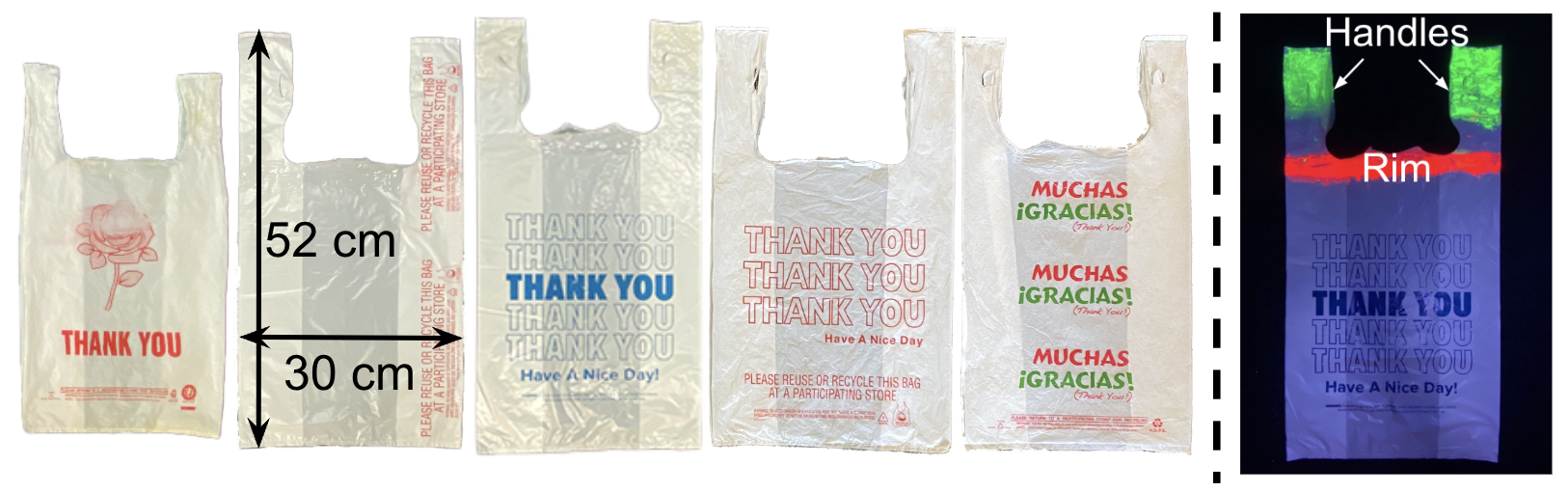}
\caption{
\textbf{Left}: 5 plastic bags. The first 4 are used to train the perception module (Sec.~\ref{subsec:perception}). Bags 1 and 5 (test bag) are used in experiments (Sec.~\ref{sec:exp}). \textbf{Right}: Bag 3 painted with green UV paint on its handles and red UV paint around its rim, under UV lighting.
}
\vspace{-15pt}
\label{fig:bags}
\end{figure}

\section{Problem Statement}\label{sec:PS}

\begin{figure*}[t]
\center
\includegraphics[width=1.00\textwidth]{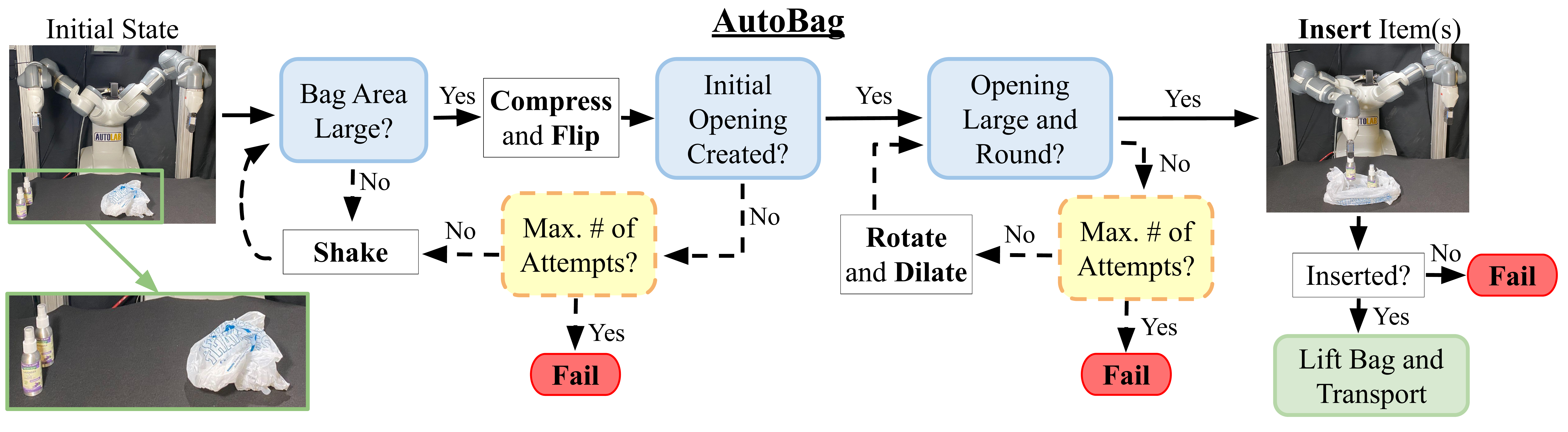}
\caption{
Overview of \bagging. The robot starts with an unstructured bag with objects, shown to the left. Given this setup, \bagging follows the procedure shown in the flow chart to first orient the bag upward and then enlarge the bag opening (see Section~\ref{subsec:algorithm} for details). If the bag opening metric exceeds a threshold, then \bagging proceeds to the item insertion stage (Section~\ref{subsec:insertion}). A trial is a full success if the robot lifts the bag with all items in it. *Not shown in the figure: In the implementation, at the initial state, the robot will also check whether there is already an initial opening so it does not unnecessarily perform a \textbf{Compress} and \textbf{Flip} and potentially ruin a good state.
}
\vspace{-10pt}
\label{fig:system}
\end{figure*}

We propose \emph{bagging}: autonomously manipulating a thin plastic bag to open it, insert at least one item, and lift it for transport.
We consider the broad class of thin plastic shopping bags commonly used in grocery stores that are made of a single sheet of translucent, reflective, and highly flexible thin plastic material cut with two holes for handles from die-cutting~\cite{wiki:die-cutting}. 
We define the bag ``rim'' as the edge of the bag around its primary opening as if the plastic handles were cut off. See Figure~\ref{fig:bags} for an illustration. The term ``opening'' in this definition refers to the planar surface enclosed by the rim, through which objects are put inside the bag. The orientation of the opening is the direction of the outward-pointing normal vector from the plane formed by the opening.
In many configurations, the opening can have an area of zero (\eg when the rim is fully folded).

We assume the initial bag state is unstructured: deformed and potentially compressed, and resting stably on the surface, in which the bag rim and handles may be partially or fully occluded.
We assume that the two sides of the bag do not stick to each other tightly (which occurs in new bags) and that some opening can be created using actions such as shaking, flinging, or pulling.
We assume that the bag can be segmented from the workspace (\eg by color thresholding). 

We consider a bimanual robot with parallel-jaw grippers and an overhead calibrated RGB camera above a flat surface.
At each timestep $t$, the robot receives an RGB image observation $I_t \in \R^{W \times H \times 3}$ of the bag configuration $\bs_t$, and executes an action $\ba_t$. We assume also a set of small rigid objects $\mO$, placed in known poses for grasping and insertion. The objective is to develop a policy $\pi: I \mapsto \ba$ so that the robot opens the bag, places at least one object inside the bag, then lifts the bag off the table while containing the objects for transport.

\section{Method}\label{sec:method}

We propose a learned perception module to recognize the bag rim and handles that includes a novel self-supervised data collection process (Section~\ref{subsec:perception}) for training, during which the robot uses its action primitives 
(Section~\ref{subsec:primitives}). 
We propose two metrics for quantifying the bag opening (Section~\ref{subsec:metric}). We then describe the \bagging algorithm (visualized in Figure~\ref{fig:system}) for opening a plastic bag 
(Section~\ref{subsec:algorithm}), followed by item insertion (Section~\ref{subsec:insertion}).

\begin{figure*}[t]
\center
\includegraphics[width=1.00\textwidth]{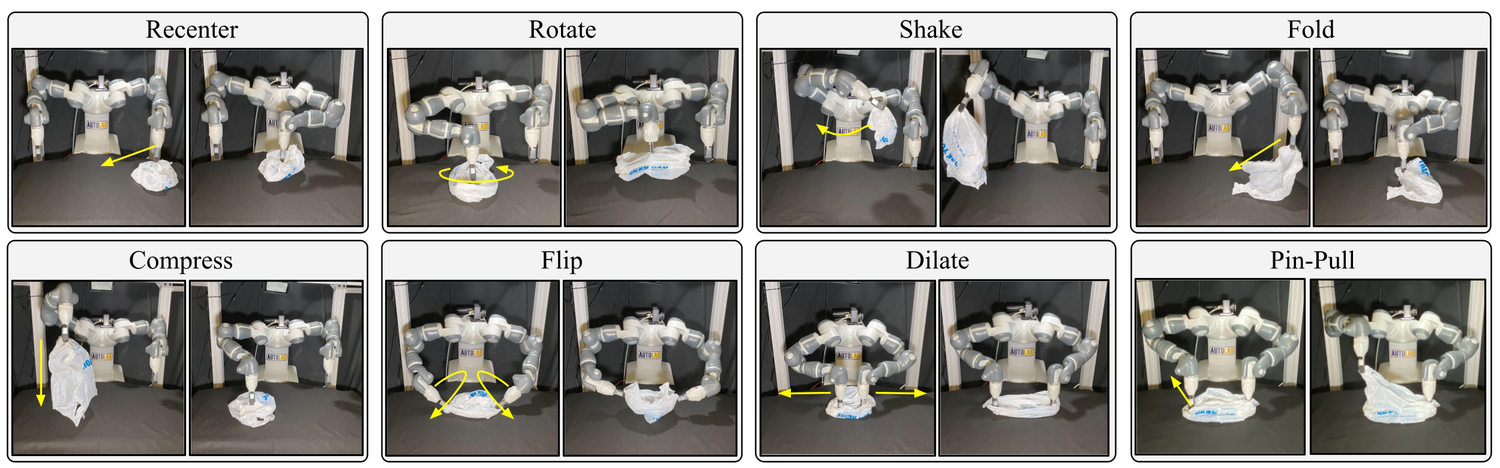}
\caption{
The action primitives in this work. See Section~\ref{subsec:primitives} for details, and Section~\ref{subsec:insertion} for \textbf{Pin-Pull} (only used during item insertion). For each primitive, we show a small two-frame overview of it in action. The project website contains full videos of all primitives.
}
\vspace{-10pt}
\label{fig:primitives}
\end{figure*}

\subsection{Self-Supervised Learning for Perception Module}
\label{subsec:perception}

In this work, we propose to represent bags through semantic segmentation. With this representation, we conjecture that the robot may be able to estimate the bag state from complex and deformed bag configurations.
The robot perceives 2 key parts of a bag:  \textbf{bag handles} and \textbf{bag rim} (see Figure~\ref{fig:bags}). We formulate this as an image segmentation task where, given an RGB image, the output is a per-pixel classification among 4 classes: the 2 aforementioned semantic regions, any remaining area of the bag, and the background.

We use a self-supervised data collection procedure for training a segmentation model using UV-fluorescent markings to avoid time-consuming and expensive human annotation~\cite{LUV_2022}.
We place 6 programmable UV LED lights overhead and paint the 2 key parts of bags (handles and rim) with transparent UV-fluorescent paints that brightly reflect 2 different colors under UV light. 
When the UV lights are turned off, the paints are invisible and the bag looks normal under regular lighting. When the regular lights are turned off and the UV lights are turned on, everything is dark except for the regions with UV paints, which glow their unique colors. 

With this setup, the robot uses its action primitives (see Section~\ref{subsec:primitives}) to manipulate the bag into different configurations, and by alternating the lighting conditions, the camera collects paired images of the bag in both standard and UV lighting.
By extracting the segmentation masks from the image under the UV lights through color thresholding, the system obtains the ground truth segmentation labels corresponding to the image of the bag under regular lighting conditions, which are then used to train the segmentation network.
The objective of this process is to manipulate a plastic bag into a diverse set of configurations---both in terms of its volume and its orientation---as the increased data diversity can lead to a higher-quality perception model.

\subsection{Action Primitives}\label{subsec:primitives}

We consider a set of action primitives $\mathcal{A}$ (see Figure~\ref{fig:primitives}). Each primitive has a type $m \in \mathcal{M}$ and action-specific parameters $\phi_m$:
$\ba = \langle m, \phi_m \rangle \in \mathcal{A}.$
Gripper positions are specified as Cartesian coordinates in pixels, and the grasping height is set to the height of the workspace to ensure successful grasping of the bag. 
The primitives are:

\begin{enumerate}[leftmargin=*]
\item \textbf{Recenter $(x, y)$}: Grasp the bag at $(x,y)$ from top down, lift it up, and translate it to the workspace center at $(0,0)$. This prevents the bag from moving off the workspace.

\item \textbf{Rotate $(x, y, \alpha, \beta, \gamma)$}: Grasp the bag at $(x,y)$ from top down, lift it up, rotate the gripper by 
Euler angles $(\alpha, \beta, \gamma)$, and then directly place the gripper down. 

\item \textbf{Shake $(x, y, k_s, \ell, f)$}: Grasp the bag at $(x, y)$ from top down, lift it up, then perform $k_s$ shaking motions of amplitude $\ell$ and frequency $f$, where the gripper rotates its wrist side by side, followed by a swing action to lay the bag on the table. This action often expands the surface area of a compressed bag.

\item \textbf{Fold $(x, y, d)$}: Grasp the bag at $(x,y)$ from top down, lift it off the workstation, move the gripper horizontally outward by distance $d$ and then inward and downward to fold the bag and reduce its top-down visible surface area.
This action enables the robot to compress and partially reset the bag state. Combined with other actions that expand the bag, we find that this induces greater diversity of bag configurations during self-supervised data collection (Section~\ref{subsec:perception}). The robot does not use this action when opening the bag at execution time (Section~\ref{subsec:algorithm}).

\item \textbf{Compress $(x, y, k_c)$}: Grasp the bag at $(x,y)$, lift it up in midair, then press it downwards until contact with the workspace, and repeat for a total of $k_c$ motions. This action changes the side of the bag that is flat after being compressed. 
Holding the bottom part with the bag opening facing downward and compressing also allow air to inflate the bag opening. 

\item \textbf{Flip $(x_l, y_l, x_r, y_r, \alpha)$}: Orient both grippers towards each other, each with an angle $\alpha$ with the horizontal plane, grasp opposite ends of the bag at positions $(x_l, y_l)$ and $(x_r, y_r)$. Then, lift both grippers, rotate each gripper by 180 degrees, and then place the bag down. This action tends to change the direction of the bag opening (\eg from downwards to upwards).

\item \textbf{Dilate $(x_l, y_l, x_r, y_r, \alpha, \theta, d)$}: Bring both grippers together at positions $(x_l, y_l)$ and $(x_r, y_r)$ and with them pointing downwards with an angle $\alpha$ with the horizontal plane. Then move both grippers away from each other horizontally along direction $\theta$, each moving by distance $d$. 
This action can enlarge a small opening.
We use this action during bag opening but not during data collection.
\end{enumerate}

During data collection, the robot uses the following policy to select actions to manipulate the bag into diverse states:
\begin{enumerate}[leftmargin=*]
    \item When the bag area (as viewed from above) exceeds a threshold, sample these primitives uniformly at random: \textbf{Rotate, Shake, Fold, Compress, Flip}; 
    \item When the bag area is small, sample these primitives uniformly at random: \textbf{Rotate, Shake, Compress};  
    \item After \textbf{Compress}, perform a \textbf{Flip};  
    \item When the bag is off the center, perform \textbf{Recenter}.
\end{enumerate}

\begin{figure*}[t]
\center
\includegraphics[width=1.00\textwidth]{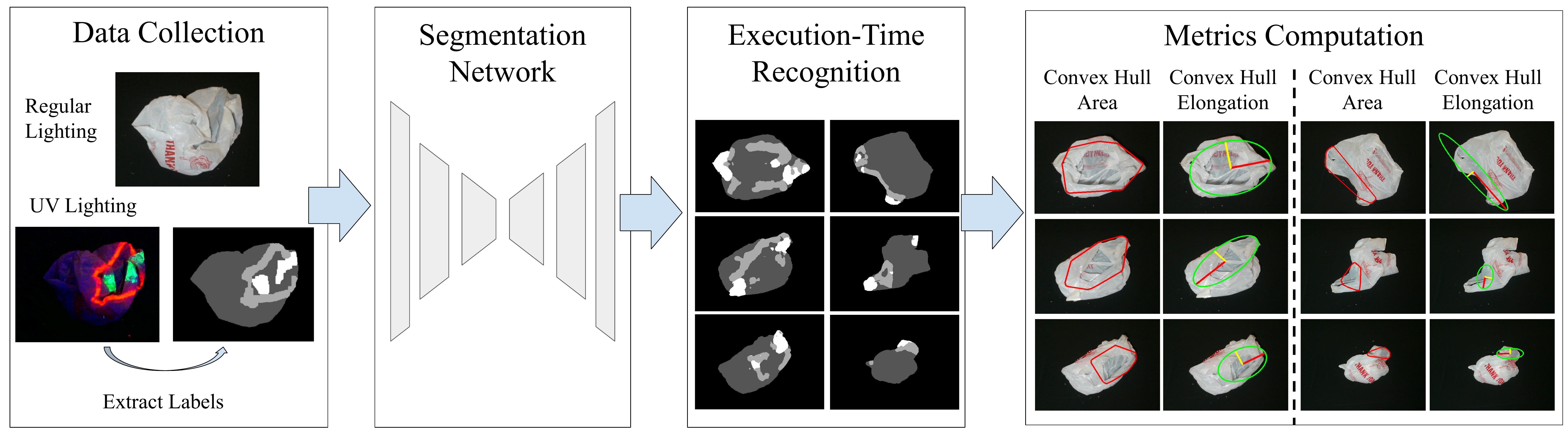}
\caption{
Perception Pipeline. The robot collects images by autonomously manipulating a bag into diverse configurations. It extracts segmentation labels of bag handles and rims through color thresholding from the images under UV light. We use this data to train a semantic segmentation network. During execution time, the perception model takes in an RGB image and predicts the segmentation (white: handles; light gray: rim; dark gray: other parts of the bag; black: background). We fit a convex hull to the predicted rim region (indicated by the red boundary), and compute the area and elongation of the convex hull to quantify the bag opening (Section~\ref{subsec:metric}). We overlay an ellipse with the major and minor axes computed by PCA on the convex hull.
}
\vspace{-10pt}
\label{fig:perception}
\end{figure*}

\subsection{Bag Opening Metrics}\label{subsec:metric}

We propose to use two metrics for quantifying the bag opening from a segmented image.
Both use the convex hull of the bag rim, denoted as $CH$, which is the convex region in the image enclosed by the pixels that the perception model (Section~\ref{subsec:perception}) identifies as the rim. 

\begin{enumerate}[leftmargin=*]
    \item \textbf{Normalized convex hull area} $A_{CH}$: to approximate the size of the bag opening, we compute its convex hull area in 2D pixel space, and divide that by the maximum convex hull of the bag rim. To compute the  maximum convex hull value, a human manually manipulates the bag offline to maximize the opening.
    \item \textbf{Convex hull elongation} $E_{CH}$: we approximate the bag opening elongation using the ratio of the PCA major and minor axes of the convex hull in 2D pixel space. 
\end{enumerate}

See Figure~\ref{fig:perception} for visualizations. The normalized convex hull area makes the metric bag size-agnostic.  In addition, we use convex hull elongation because we observe that for inserting items, a sideways-facing bag with a closed opening is worse than an upward-facing opening which is small but rounded. The normalized convex hull area, however, may give higher values to the former. Consequently, measuring elongation gives extra information about the bag opening.

\subsection{\bagging Algorithm}\label{subsec:algorithm}
The first part of the \bagging algorithm uses the perception module (Section~\ref{subsec:perception}) to choose actions (Section~\ref{subsec:primitives}) to open a bag from an unstructured initial state. See Figure~\ref{fig:system} for an overview. This consists of the following two stages.

\textbf{\underline{Stage 1}}:  If the 2D pixel surface area of the bag is below a threshold $S^{(1)}$, the robot executes a \textbf{Shake} to expand the bag, where its grasp points are sampled from the handle region (or anywhere on the bag if handles are not visible). Otherwise, if the area exceeds $S^{(1)}$, the robot grasps the bottom and executes a \textbf{Compress}. This flattens the bottom so that when the robot next executes a \textbf{Flip}, the bag may stand stably with its opening facing upward. With an updated top-down image, the algorithm uses the perception model to compute the metrics $A_{CH}$ and $E_{CH}$. When the former is above threshold $A_{CH}^{(1)}$ and the latter is below threshold $E_{CH}^{(1)}$, it is likely that an initial upward-oriented opening exists. If the thresholds are not met, the bag is likely tilted on its side or folded inward, with the opening closed, and the robot resets the state by executing a \textbf{Shake} and repeats.

\textbf{\underline{Stage 2}}:  The robot uses \textbf{Rotate} and \textbf{Dilate} actions to iteratively enlarge the opening. At each iteration, the algorithm queries an overhead image of the bag, estimates its rim positions, and uses PCA to identify the direction of the major and minor axes of the opening. The robot performs a \textbf{Rotate} about the $z$-axis so that the minor axis of the bag aligns with the horizontal axis, and then uses \textbf{Dilate} to pull the bag opening farther apart from the opening center along its minor axes.
This process repeats until the normalized convex hull area reaches a threshold $A_{CH}^{(2)}$ and the elongation metric falls below a threshold $E_{CH}^{(2)}$, suggesting that the opening is large and round enough for object insertion (Section~\ref{subsec:insertion}).

\subsection{\bagging: Object Insertion and Bag Lifting}\label{subsec:insertion}
The final steps of \bagging involve inserting the objects into the bag and lifting the bag. To lift the bag, the robot grasps the bag at the workstation height to avoid missed grasps, but this can lead to grasping multiple layers, a common challenge in deformable manipulation~\cite{tirumala2022}. 
We thus propose the \textbf{Pin-Pull $(x_{pin}, y_{pin}, x_{pull}, y_{pull})$} primitive: one gripper goes to position $(x_{pin}, y_{pin})$, and presses down onto the bag (``pinning''). The other gripper goes to position $(x_{pull}, y_{pull})$, closes the gripper, and lifts up to a fixed height $h$ or until a torque limit is reached (``pulling''). 
The purpose of \textbf{Pin-Pull} is to stretch the bag after grasping. The additional layer that is accidentally grasped can slip out of the gripper during this process, leading to a higher success rate of grasping a single layer.

Given an overhead image of the bag, the robot estimates the opening by fitting a convex hull on the perceived rim, then divides the space by the number of objects. It grasps each object using known poses and places them in the center of each divided region. Then it identifies the positions of the handles and performs two \textbf{Pin-Pull} actions to grasp the left and right handles (or bag boundaries if handles are occluded). Finally, the robot lifts the bag off the table.

\section{Physical Experiments}\label{sec:exp}

During training and experiments,
the bags we use are of size \SI{28}-\SI{30}{\centi\meter} by \SI{49}-\SI{54}{\centi\meter} when laid flat (see Figure~\ref{fig:bags}). The flat workspace has dimensions \SI{70}{\centi\meter} by \SI{90}{\centi\meter}. 

\subsection{Self-Supervised Training of Perception Module}\label{subsec:training}

For data collection, we use 4 training bags (see Figure~\ref{fig:bags}) and collect 500 images for each bag. 
All images come with automatic labels using the self-supervised procedure. 
We use a U-Net architecture~\cite{ronneberger2015u} for the segmentation network, trained with soft DICE loss~\cite{milletari2016v}. 
We use one NVIDIA Titan Xp GPU, with a batch size of 8, and a learning rate of 5e-4. 
The trained model achieves a 77\% intersection over union (IOU) on the validation set. 
See the supplement for the learning curve and example predictions.

\subsection{Experiment Protocol}\label{subsec:protocol}

To evaluate \bagging, we use two bags, one of which is the smallest bag from training. The other, unseen bag has the same size as the largest training bag but has different patterns (see both in Figure~\ref{fig:bags}). Neither has UV paint.
The goal is to insert 2 identical 2 oz. spray bottles into each bag. 
We define a \emph{trial} as an instance of the robot attempting to perform the full end-to-end procedure: to open a bag, insert the items into it, and then lift the bag (with items) off the surface. We allow for up to $T=15$ actions (excluding \textbf{Recenter}) before the robot must formally lift the bag. If the robot encounters motion planning or kinematic errors during the trial, we reset the robot to the home position and continue the trial.
We consider 3 tiers of initial bag configurations:
\begin{itemize}[leftmargin=*]
    \item \underline{Tier 1}: The bag starts upward-facing with the rim recognizable but with a small opening. This requires enlarging the bag opening to allow placing objects inside. 
    \item \underline{Tier 2}: The bag starts at an expanded, slightly wrinkled state lying sideways on the workspace. This requires reorienting the bag upwards and then opening the bag. 
    \item \underline{Tier 3}: Any other, more complex configuration, such as when the bag is compressed with no visible handles. 
\end{itemize}

At the start of each trial, we initialize the bag by compressing and then adjusting it to make it fall into one of the tiers. A trial is an ``$n\geq1$ success'' if the robot can contain at least one bottle in the bag when lifting. A trial is an ``$n=2$ success'' when the robot lifts the bag off the surface while containing both bottles. Additionally, we report the number of times the robot successfully opens the bag (``Open Bag'') (i.e., proceeds to item insertion), and the number of objects that the robot correctly places in the bag opening before lifting (``\#Placed''). Objects may fall out when the robot lifts the bag, meaning that the number of objects contained (``\#Contained'') is upper-bounded by \#Placed.

\subsection{\bagging and Baseline Methods}\label{subsec:baselines}

We compare \bagging to baselines and ablations, which we summarize here and defer details to the supplement.

To evaluate the proposed perception module, we perform ablation studies in Tier 1 and Tier 2 that remove the perception module (``AB-P''). Instead of detecting the rim and using the rim area and elongation as the metrics for determining whether the bag is sufficiently opened, it uses the area and elongation of the bag as an approximation. In addition, for \textbf{Dilate}, instead of dilating from the opening center, it uses the bag center. To evaluate the proposed metrics for quantifying the bag opening, we perform ablation studies in Tier 1. ``AB-$A_{CH}$'' uses only the convex hull elongation metric, and ``AB-$E_{CH}$'' uses only the convex hull area metric.

Additionally, for Tier 2, since the bag opening faces sideways, we design 2 different baselines. 
``Side Ins.'' attempts to grasp the top layer of the bag to open it and then insert objects from the side. ``Handles'' grasps the two handles, performs horizontal and vertical flinging to open the bag, and then releases one gripper to insert the objects while the other gripper still holds the bag.

\section{Results and Limitations}\label{sec:results}

\begin{table}[t]
  \setlength\tabcolsep{5.0pt}
  \centering
  \vspace{-0em}
  \vspace{1pt}
  \centering
  \footnotesize
  \begin{tabular}{ccccccc}
  \toprule
  Train & \multirow{2}{*}{Method} & Open & \multirow{2}{*}{\#Placed} & \multirow{2}{*}{\#Contained} & $n \geq 1$ & $n = 2$ \\
  Bag   &        & Bag  &          &             & Succ. & Succ. \\
  \midrule
  \multirow{3}{*}{Tier 1} & AB-P & 2/6 & 0.7$\pm$1.0 & 0.7$\pm$1.0 & 2/6 & 2/6 \\
   & AB-$A_{CH}$ & 5/6 & 1.0$\pm$0.9 & 0.5$\pm$0.5 & 3/6 & 0/6  \\
   & AB-$E_{CH}$ & 5/6 & 0.8$\pm$1.0 & 0.7$\pm$1.0 & 2/6 & 2/6  \\
   & \bagging   & \textbf{5/6} & \textbf{1.6$\pm$0.7} & \textbf{1.1$\pm$0.8} & \textbf{4/6} & \textbf{3/6} \\
  \midrule
  \multirow{3}{*}{Tier 2} & Side Ins. & 1/6 & 0.3$\pm$0.8 & 0.3$\pm$0.8 & 1/6 & 1/6 \\
   & Handles & 2/6 & 0.2$\pm$0.4 & 0.2$\pm$0.4 & 1/6 & 0/6 \\
   & AB-P & 2/6 & 0.7$\pm$1.0 & 0.5$\pm$0.8 & 2/6 & 1/6 \\
   & \bagging   & \textbf{4/6} & \textbf{1.3$\pm$1.0} & \textbf{0.8$\pm$1.0} & \textbf{4/6} & \textbf{2/6} \\
  \midrule
  Tier 3 & \bagging   & \textbf{1/6} & \textbf{0.3$\pm$0.8} & \textbf{0.3$\pm$0.8} & \textbf{1/6} & \textbf{1/6} \\
  \bottomrule
  \end{tabular}
  \caption{
  Physical experiment results of \bagging, ablations, and baseline methods using a training bag across 3 different tiers.
  }
  \label{tab:tab1}
  \vspace*{-15pt}
\end{table}

\begin{table}[t]
  \setlength\tabcolsep{5.0pt}
  \centering
  \vspace{-0em}
  \vspace{1pt}
  \centering
  \footnotesize
  \begin{tabular}{lccccc}
  \toprule
  \bagging & Open & \multirow{2}{*}{\#Placed} & \multirow{2}{*}{\#Contained} & $n \geq 1$ & $n = 2$ \\
  Test Bag & Bag  & & & Succ. & Succ. \\
  \midrule
  Tier 1 & 4/6 & 1.2$\pm$1.0 & 1.0$\pm$0.9 & 4/6 &  2/6 \\
  Tier 2 & 3/6 & 0.9$\pm$1.0 & 0.7$\pm$0.8 & 3/6 & 1/6 \\
  \bottomrule
  \end{tabular}
  \caption{
  Results of \bagging on the test bag, across the first two tiers. We report similar evaluation metrics as in Table~\ref{tab:tab1}.
  }
  \label{tab:tab2}
  \vspace*{-15pt}
\end{table}

We report results on the training bag in Table~\ref{tab:tab1}, where we run 6 trials per experiment setting. 
The results suggest that \bagging achieves an $n=2$ success rate of 3/6, 2/6, and 1/6 on the three respective tiers and outperforms baselines and ablations in all metrics. The low success rate of ``AB-P'' for opening the bag suggests the perception module is important for determining whether a bag is actually opened and for placing the gripper inside the opening for \textbf{Dilate}. Using only the opening area metric achieves a good $n \geq 1$ success rate but fails to put both objects in, as the opening is not sufficiently wide. On the other hand, using only the elongation metric ensures the bag opening is round but does not guarantee that the opening is large enough.
For sideways insertion, the main challenge is grasping the top layer only to create a sideways opening. We find that it often either misses the grasp or grasps two layers, due to the tight space between the two layers. The main failure reason for the ``Handles'' baseline is that the bag collapses or the opening closes as soon as one gripper releases the handle to grasp the inserted object, leaving no room for the robot to insert the objects.

On the test bag, we conduct experiments on Tier 1 and Tier 2 only, due to the extra challenge from the test bag perception. The results in Table~\ref{tab:tab2} suggest that \bagging can attain 4/6 and 3/6 partial successes.

We summarize the failure modes of \bagging: 
\begin{enumerate}[(A)]
    \item Number of trials reaches the maximum limit (38\%).
    \item Bag gets pushed out of the workspace (3\%).
    \item Objects are not placed inside the opening (13\%).
    \item Objects falls out when lifting the bag (23\%).
    \item Fail to lift the bag successfully (23\%).
\end{enumerate}

(A) is often caused by instability of the perception model prediction, especially in complex bag configurations and around decision boundaries, leading to unsuitable/ineffective actions. 
(B) occurs due to robot arm motion planning. (C) is due to inaccuracy in the rim recognition, especially for the unseen test bag. 
(D) is a major bottleneck for achieving \emph{full} success, 
due to non-ideal grasp position for lifting. (E) is caused by slipped grasps. More details, including failure reasons for baselines, are in the supplementary material.

\section{Conclusion and Future Work}

In this paper, we propose a novel problem and an algorithm, \bagging, for opening a thin plastic bag and inserting items. In future work, we will study better manipulation techniques 
to increase the success rate and speed up \bagging.

\section*{Acknowledgements}
{\footnotesize
This research was performed at the AUTOLAB at UC Berkeley in affiliation with the Berkeley AI Research (BAIR) Lab, and the CITRIS ``People and Robots'' (CPAR) Initiative. The authors were supported in part by donations from Toyota Research Institute.  Lawrence Yunliang Chen is supported by the National Science Foundation (NSF) Graduate Research Fellowship Program under Grant No. 2146752. Daniel Seita and David Held are supported by NSF CAREER grant IIS-2046491. We thank Justin Kerr, Roy Lin, and the reviewers for their valuable feedback.}

\renewcommand*{\bibfont}{\footnotesize}
\printbibliography

\clearpage
\section{Appendix}

\subsection{Action Primitives Implementation Details}
Here, we describe the implementation details for the action primitives.
\begin{itemize}
    \item For \textbf{Shake}, we set number of repetitions $k_s = 3$, 
    amplitude $l = 0.7\pi$ radian ($40.1\degree$), and frequency $f = 0.4$~Hz. 
    \item For \textbf{Fold}, we set $d = 28$~cm. 
    \item For \textbf{Compress}, we set $k_c = 4$, with a pause of $0.9s$ after every repetition to let the bag settle. The robot grasps the bag with an angle $\alpha = \pi / 7$ radian ($25.7\degree$) with the horizontal plane.  
    \item For \textbf{Flip}, we set the gripper grasping angle $\alpha = 45\degree$.
    \item For \textbf{Dilate}, we restrict the dilation angle $\theta = 0 \degree$ and set distance $d = 12$ cm. Gripper stops when it has moved by $d$ or the torque sensor has reached threshold of $0.05\si{\newton\meter}$.
\end{itemize}

\subsection{Details of Grasp Point Selection}
During data collection, the grasp points for each primitive are sampled as follows:
\begin{itemize}
    \item \textbf{Rotate}: Uniformly on the bag.
    \item \textbf{Shake}: Uniformly from the boundary of the bag.
    \item \textbf{Fold}: Uniformly from the boundary of the bag.
    \item \textbf{Compress}: Uniformly from the bag.
    \item \textbf{Flip}: The left and right endpoints of the bag segmentation along a horizontal line.
    \item \textbf{Recenter}: Center of the bag segmentation. 
\end{itemize}

During the execution of \bagging, the grasp points for each primitive are sampled as follows:

\textbf{\underline{Stage 1}}: 
\begin{itemize}
    \item \textbf{Shake}: Grasp the center of the handle if at least one handle is visible. Otherwise, sample uniformly from the boundary of the bag.
    \item \textbf{Compress}: The robot grasps the bag with an angle $\alpha = \pi / 7$ radian ($25.7\degree$) with the horizontal plane. The grasp point is the center of the bottom region of the bag, where the bottom is inferred from the rim of the bag using a heuristic: Fit a rectangle to the bag, and find the two corners that are farther from the rim. The center bottom of the bag is then approximated by finding the midpoint of those two farther corners and shrink towards the bag center until the midpoint lies on the bag.
    \item \textbf{Flip}: Same as during data collection.
\end{itemize}

\textbf{\underline{Stage 2}}: 
\begin{itemize}
    \item \textbf{Rotate}: Center of the bag segmentation.
    \item \textbf{Dilate}: The two grippers are oriented towards each other with angle $\alpha = \pi / 3$ radian ($60\degree$) with the horizontal plane and positioned around the center of the bag opening with offset by $0.02$ cm each to avoid collision. The bag opening is approximated using the convex hull of the rim prediction. $\theta$ is set to be $0\degree$, and $d = 10$ cm. Gripper stops when it has moved by $d$ or the torque sensor has reached threshold of $0.02\si{\newton\meter}$.
\end{itemize}

\textbf{\underline{Bag Lifting}}: 
\begin{itemize}
    \item \textbf{Pin-Pull}: The two gripper positions $(x_{pin}, y_{pin})$ and $(x_{pull}, y_{pull})$ are the center of the two predicted handles if the handles are visible and the left and right endpoints of the bag segmentation if the handles are not visible.
\end{itemize}

\subsection{Details of Perception Model}\label{app:perception}
\begin{figure}[t]
\center
\includegraphics[width=0.23\textwidth]{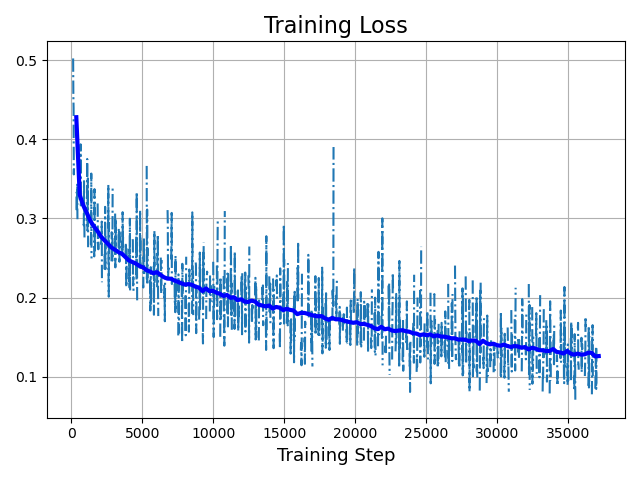}
\includegraphics[width=0.23\textwidth]{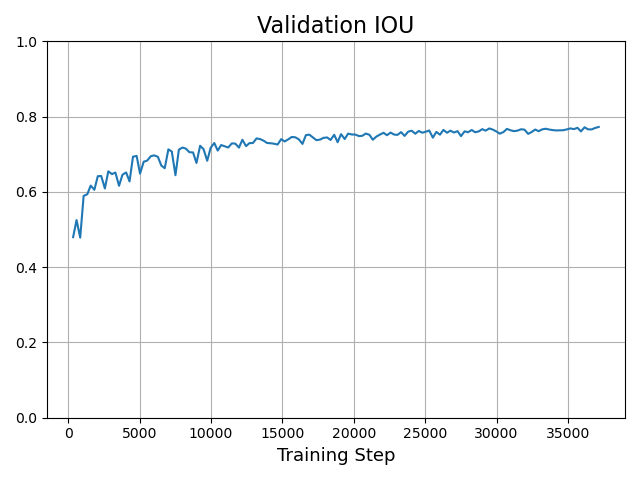}
\caption{
Learning curves of the perception model (Section~\ref{subsec:training}). \textbf{Left}: Training loss. \textbf{Right}: IOU on the validation set.
}
\label{fig:learning_curve}
\end{figure}

We collect 2,000 paired regular-UV images in total across 4 training bags. Collecting each regular-UV image pair takes 30-40 seconds. To obtain the ground truth segmentation, we use color thresholding from the UV images and dilate the rim and handle labels to connect regions where the UV paint does not glow strongly and filter noises. We use an 80-20 train/validation split for model training. We use one NVIDIA Titan Xp GPU, with a batch size of 8, a learning rate of 5e-4, and a weight decay factor of 1e-6.  Figure~\ref{fig:learning_curve} shows the learning curve, including the training loss and validation intersection over union (IOU).

\begin{figure}[t]
\center
\includegraphics[width=0.49\textwidth]{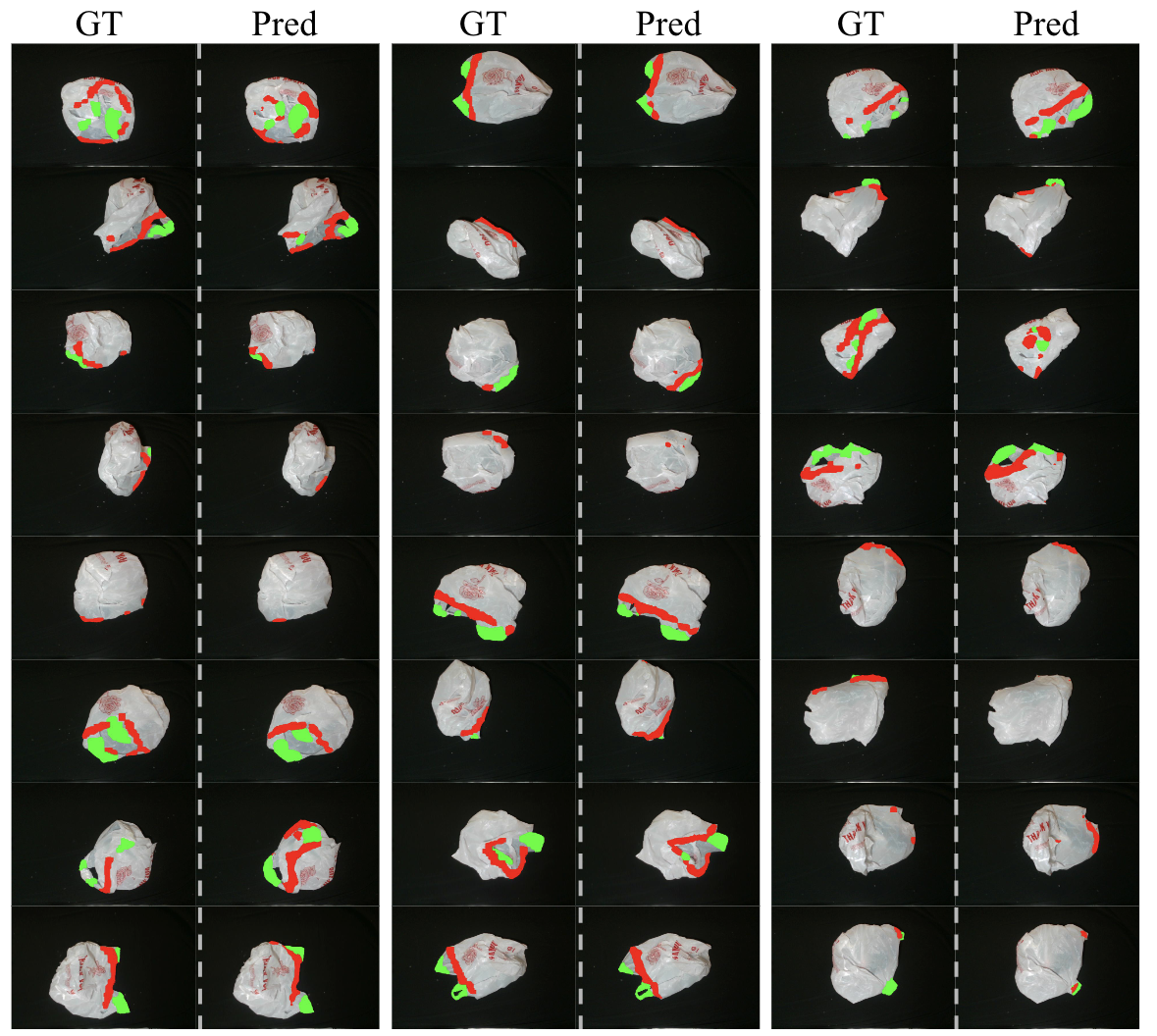}
\caption{
Examples of the prediction of the perception model on the validation dataset. For each of the three columns, the left ``GT'' are the ground truth labels, and the right ``Pred'' are the model predictions. Both are overlaid on top of the color images, with green indicating handles and red indicating rim.
}
\label{fig:example_prediction}
\end{figure}

\begin{figure*}[t]
\center
\includegraphics[width=1.00\textwidth]{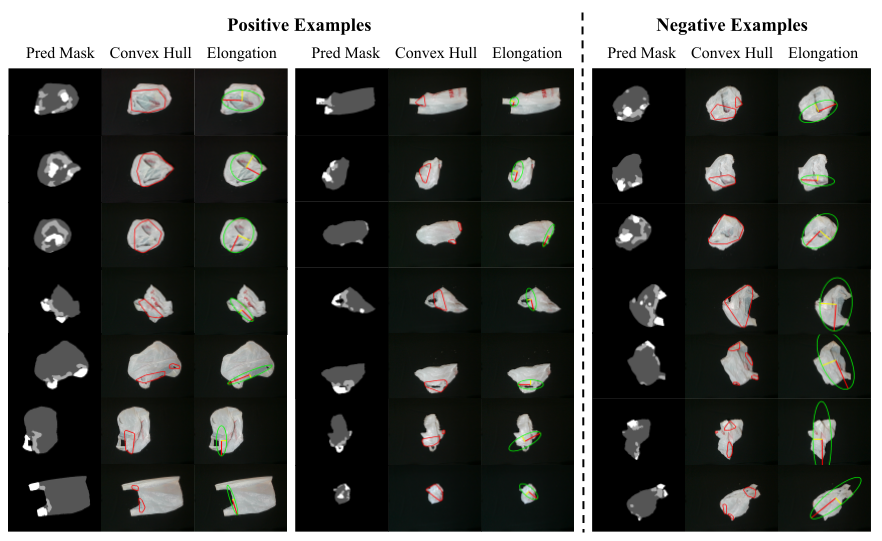}
\caption{
Positive and negative example predictions of bag segmentation and illustrations of the metrics (convex hull and elongation) applied to the predicted rim.
}
\label{fig:metrics_pos_neg_examples}
\end{figure*}

Figure~\ref{fig:metrics_pos_neg_examples} illustrates some positive and negative example predictions of bag segmentation and the metrics (convex hull and elongation) applied to the predicted rim. We can see that negative examples are mainly due to the wrong predicted rim mask, including both type 1 and type 2 errors. This can lead to the estimated opening through the convex hull being either too large (including other parts of the bag) or too small (when only part of the true rim is recognized). When the predicted rim is noisy and segmented all over the bag, this can also lead to an inaccurate elongation estimate.

\subsection{Experiment Details}
For \textbf{\bagging}, we define $S^{(1)}$ to be the bag size in 2D pixel space divided by the maximum bag size obtained from a flat bag and
we set $S^{(1)} = 0.55$, $A_{CH}^{(1)} = 0.15$, $E_{CH}^{(1)} = 4.5$, $A_{CH}^{(2)} = 0.45$, $E_{CH}^{(2)} = 2.88$.

For the ``Side Ind.'' baseline, we apply the \textbf{Pin-Pull} primitive to the opening of the bag. In particular, we choose the pin position $(x_{pin}, y_{pin})$ as the center of the rim farther from the bag center, and the pull position $(x_{pull}, y_{pull})$ as the midpoint between the center of the rim closer to the bag center and the bag center. The intention is to pin the bottom layer of the bag while pulling the top layer of the bag in order to separate the two layers of the bag and create a sideways-facing opening. The robot then attempts to insert the object from the side, regardless of whether the two layers have been truly separated or not, since the overhead camera cannot tell this.

For the ``Handles'' baseline, the robot first grasps the two handles, one in each gripper. It then lifts the bag in midair and performs two sequences of dynamic actions. It first shakes the bag in the horizontal left-right direction 2 times. This has the effect of separating the two layers of the rim to prevent them from sticking to each other. The robot then flings the bag vertically 3 times. This action allows the air to come into the bag opening and inflate the bag. Then, the robot’s right gripper releases the handle to grasp the object. 
Finally, the robot inserts the object at the center of the estimated bag opening while the other gripper still holds the bag.

\subsection{Failure Modes for \bagging}
Here, we describe the details of the failure cases for \bagging and baselines.

For \bagging, we divide the failure modes into 5 categories. 

(A) is caused by both manipulation and perception challenges during the long sequences of manipulation steps. In particular, during manipulation, the gripper may miss the grasp if the grasp region is too flat on the surface to have enough friction, or the bag may slip out of the gripper during dynamic actions such as \textbf{Shake} and \textbf{Compress}. Additionally, the perception module is not always robust and may sometimes only recognize part of the rim. This leads to the convex hull approximation of the opening underestimating the true opening and causes the robot to perform more Stage 1 and Stage 2 actions than necessary. For example, after \textbf{Compress} and \textbf{Flip}, even when the bag has a large opening, the robot may fail to recognize the entire rim region, leading to the metrics not meeting the thresholds for Stage 2 and causing the robot to repeat the Stage 1 actions. Another example of unnecessary actions can happen during \textbf{Rotate} and \textbf{Dilate}, where the perception module only recognizes part of the opening, causing the robot to keep rotating and dilating the bag. While this conservatism is not fatal on its own, the increased action steps lead to a larger chance of failure rate due to imperfect manipulation, as the opening can easily close again during manipulation. For example, during \textbf{Dilate}, the friction of the bag with the two grippers may not be the same, causing the bag to slide along with one gripper while the other gripper slips. This moves the bag off the center of the workspace. Also, \textbf{Dilate} is effective only when the two grippers start inside or close to the opening. When the perception module fails to recognize the entire opening, the estimated center of the opening will be inaccurate, causing one gripper to start inside the opening and the other gripper to start far from the opening. When sliding in this asymmetric configuration, the grippers not only fail to enlarge the opening but also tend to compress and realign the separating rim of the bag, making it more difficult for future \textbf{Dilate} actions to enlarge the bag.

(B) is caused by the motion planning of the robot. On one hand, YuMi has a limited workspace and collision-free path planning of the two arms is not always successful, especially around singularities. On the other hand, it is difficult to incorporate the bag as an obstacle for motion planning due to inaccurate depth perception and protruding loop handles. As such, the gripper may occasionally hit the bag before and after each manipulation step by accident, pushing the bag out of the workspace.

(C) is caused by inaccuracy in the rim prediction. When the perception module’s prediction of the rim has either Type 1 or Type 2 errors, the estimated convex hull will deviate from the true opening. In such cases, some objects may not be placed inside the true opening.

(D) is most often caused by non-ideal grasp positions when lifting the bag. In particular, the bag handles are often occluded by the inserted objects or hidden inside the opening after \textbf{Dilate}. In such cases, the robot simply chooses the left and right endpoints of the bag boundary. However, these grasp points may be near the bottom of the bag and lifting at those positions will make the bag upside down, so objects already inside the bag will fall out again. As the objects can all be contained inside the bag only when both grasp points are near the handles or the rim of the bag, poorly chosen grasp points of at least one gripper will prevent full success. 

(E) is caused by slipped grasps when the bag contains the inserted objects and has nontrivial weights. This is more due to the mechanical limitations of the YuMi robot, it has a limited ($500$ g) payload, not-so-strong gripping force, and poorly designed jaw grippers that will slightly tilt and open at the end when a large force is applied at the top of the metal clips to close them.

On the test bag, the perception module is the main bottleneck since it has never seen the bag before, leading to many cases where it only recognizes part but not all of the rim. As such, (A) is the dominant reason for failure.

\subsection{Failure Modes for Baselines}
Next, we describe the failure modes of the baselines.

\textbf{AB-P}: Without using the perception module, the \textbf{Dilate} action often fails to position both grippers inside the opening. As mentioned earlier, when one gripper is inside the opening while the other is outside, the \textbf{Dilate} actions can inadvertently close the bag opening due to asymmetric forces. In addition, without the rim area and elongation estimation, the robot cannot distinguish between a large bag and a large open bag, leading to inappropriate insertion actions even when the bag is not open.

\textbf{AB-$A_{CH}$}: Without using the size of the opening as a criterion for readiness to insert objects, the robot often prematurely attempts to insert objects when the opening is small but round. As a result, it is able to successfully insert 1 bottle half of the time but fails to insert both bottles in all trials.

\textbf{AB-$E_{CH}$}: Without using the roundness of the opening as a criterion, the robot would sometimes stop dilation even though the opening is very narrow. In those cases, the robot will fail to insert the objects inside the opening or lift the bag up with the objects contained since there is very little room for error.

\textbf{Side Ins.}: We observe that, unless there is already a large separation between the two layers of the bag (distance-wise) so that the pinning gripper can firmly and precisely press only the bottom layer, pulling the top layer of the bag almost always grasps the two layers together simultaneously and fails to create a sideways opening. As the layer of the plastic bag is quite thin, the pinning hand needs to exert a strong force towards the bottom layer against the table but not touch the upper layer, a stringent initial condition that is almost never satisfied with normal sideways-facing bags.

\textbf{Handles}: Failure cases can be categorized into two main reasons: (1) The fling actions fail to open the bag, and (2) the opening closes and the bag folds on itself after one gripper releases the handle. In particular, (1) is the main reason. For (1), we observe that the fling actions can only open the bag when both grippers grasp one layer of the handles (insert into the ring regions of handles). When a gripper grasps both layers of the handle, it effectively clamps the two layers of the bag together, making it difficult for the fling action to inflate the opening. For (2), we observe that even if flinging successfully inflates the bag, the opening closes as soon as one gripper releases the handle to grasp the inserted object. This is because the handle is heavy and when released, it collapses onto the bag, causing the entire bag to fold itself and the opening to close. As such, there is little room for the robot to insert the objects. Inserting the objects along the angle of the tilted opening could be helpful, but this may require more than an overhead camera to make the insertion direction very precise.

\end{document}